\documentclass[sigconf,nonacm]{acmart}

\usepackage{booktabs}
\usepackage{tabularx}
\usepackage{array}
\usepackage{multirow}
\usepackage{graphicx}
\usepackage{capt-of}
\usepackage{enumitem}

\usepackage[normalem]{ulem}

\AtBeginDocument{%
}

\begin{document}

\title{RuleForge: Inferring and Instantiating Latent Visual Rules as Executable Programs}

\author{Weifeng Xu}
\email{mc56486@um.edu.mo}
\affiliation{%
  \institution{Institute of Artificial Intelligence and Brain Sciences, University of Macau}
  \city{Macau}
  \country{China}
}
\affiliation{%
  \institution{Faculty of Information Science and Computing, University of Macau}
  \city{Macau}
  \country{China}
}

\author{Yaxiong Wang}
\email{wangyx15@stu.xjtu.edu.cn}
\affiliation{%
  \institution{Hefei University of Technology}
  \city{Hefei}
  \country{China}
}

% Corresponding author
\author{Zhedong Zheng}
\correspondingauthor
\email{zhedongzheng@um.edu.mo}
\affiliation{%
  \institution{Institute of Artificial Intelligence and Brain Sciences, University of Macau}
  \city{Macau}
  \country{China}
}
\affiliation{%
  \institution{Faculty of Information Science and Computing, University of Macau}
  \city{Macau}
  \country{China}
}

\renewcommand{\shortauthors}{Xu et al.}

\begin{abstract}
Visual rule reasoning asks whether multimodal models can recover an abstract
concept from a few examples. Existing studies either infer rules from visual
examples but evaluate only classification or verbalization, or supervise
executable generation with rules provided as input. Consequently, the
relationship between visual rule understanding and reliable rule instantiation
remains insufficiently explored. We present Bongard-Forge100k, a controlled
benchmark, together with RuleForge, a backbone-agnostic framework that jointly
learns visual rule understanding and rule-conditioned executable generation.
The benchmark contains 108{,}797 problems: 86{,}000 for training, 19{,}297 for
testing familiar rules, and 3{,}500 for testing held-out rule compositions.
The framework progresses through rule-understanding training, joint
understanding and scene-program supervision, and refinement via
execution-grounded reinforcement learning with Understanding Replay.
Executing generated programs under multiple random seeds provides verifiable
feedback on executability, rule satisfaction, diversity, and robustness. With
UniLIP as the primary backbone, RuleForge improves multiple-choice accuracy
over the base model by 20.93 percentage points on familiar rules and 28.42
percentage points on held-out rule compositions, while increasing scene-level
rule satisfaction on familiar rules from 2.91\% to 70.99\%. Additional
experiments on InternVL3.5, Qwen3-VL, and Show-o2 demonstrate the effectiveness
of the proposed training data and framework across diverse multimodal
architectures.
\end{abstract}



\begin{teaserfigure}
    \centering
    \vspace{-.1in}
    \includegraphics[width=\textwidth]{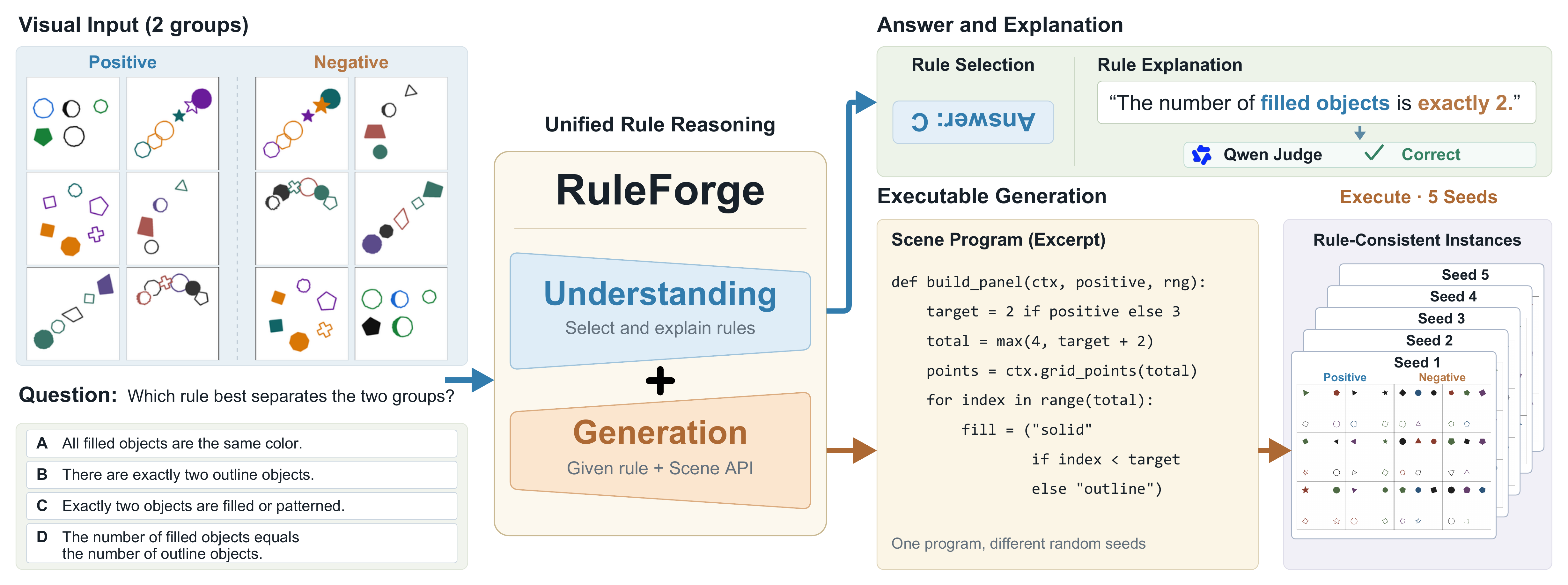}
    \vspace{-.25in} 
    \caption{\textbf{The proposed RuleForge framework for jointly learning visual rule understanding and executable generation.} Given a Bongard-Forge100k problem containing six positive and six negative panels, the model selects the separating rule and provides a natural-language explanation. Through a separate generation interface, the corresponding target rule is provided as input, and the model generates an executable scene program that produces new rule-consistent instances under five fixed seeds.\\
\emph{``What I cannot create, I do not understand.''} --- Richard Feynman
}
    \label{fig:teaser}
\end{teaserfigure}

\maketitle
\vspace{-.1in} 
\noindent\textbf{Web Relevance.}
This work advances verifiable understanding and generation of structured multimodal Web content, aligning with Web Mining and Multimedia Content Analysis.

\section{Introduction}
\label{sec:introduction}

Multimodal visual assets on the Web \cite{lin2024towards,lin2025ask}---such as charts \cite{li20252d}, infographics \cite{mathew2022infographicvqa}, and
template-driven visual layouts \cite{xie2026vsal,lin2023layoutprompter}---frequently convey meaning through structured
visual relations \cite{dai2025graph} rather than isolated appearances~\cite{
masry2022chartqa,
mathew2022infographicvqa,
lin2023layoutprompter}.
For many such assets, correctness depends in part on relational invariants:
when objects vary in scale, position, or styling, specified constraints such
as cardinality, containment, spatial relations, or symmetry should remain
satisfied. Analyzing and constructing such rule-governed visual content
therefore involves two complementary capabilities: inferring latent relational
rules from positive and negative visual examples, and constructing new
instances that verifiably satisfy explicitly specified relational constraints.

Prior visual reasoning research has largely decoupled rule understanding from
executable generation. Classical Bongard problems require a solver to infer
and verbalize the rule separating two sets of visual examples~\cite{
mikhail1970bongard,
depeweg2024solving}.
Modern Bongard-style benchmarks commonly operationalize this ability through
query-image classification, while recent studies of multimodal large language
models also examine natural-language rule descriptions~\cite{
nie2020bongard,
jiang2022bongard,
malkinski2024reasoning}.
Although these protocols evaluate whether a model can recognize or describe an
abstract regularity, they do not test whether it can instantiate an explicitly
specified rule through executable scene generation. Conversely, representative
program-aided reasoning methods generate executable intermediate programs from
explicit questions or instructions~\cite{
gao2023pal,
gupta2023visual,
suris2023vipergpt},
rather than studying whether visual rule induction and rule-conditioned scene
generation can be jointly supported by the same model. Existing protocols
therefore leave an important question unresolved: whether a shared multimodal
model can jointly acquire these two capabilities while retaining visual rule
understanding during generation-oriented optimization.

Addressing this question requires an environment in which rules are formally
defined, visual interventions are strictly controlled, and synthesized
instances can be deterministically audited. Such a benchmark must jointly
provide explicit relational rules, executable scene specifications, controlled
visual interventions, and deterministic validators for newly generated
instances. Bongard-Forge100k is designed to provide these capabilities. It
contains 108{,}797 visual reasoning problems covering 62 registered base rules
and 24 held-out rule compositions. The benchmark combines positive and
negative visual panels under systematic variations of non-target attributes
with structured rule specifications, programmatic scene generators, and
deterministic rule checkers. This design enables rigorous evaluation of
robustness to non-target visual variations under familiar rules, as well as
systematic compositional generalization to unseen rule combinations.

Building on Bongard-Forge100k, we propose RuleForge, a three-stage,
backbone-agnostic framework that progressively learns visual rule understanding
and rule-conditioned program synthesis within a shared vision-language model.
Stage~I establishes visual rule understanding through multiple-choice rule
prediction, natural-language rule explanation, candidate-rule verification,
and structured scene-fact prediction. Stage~II adds program synthesis
conditioned on separately provided target-rule descriptions and jointly
optimizes the two task families through shared model parameters and a 2:1
understanding-to-generation mixture. Stage~III refines candidate scene programs
using Group Relative Policy Optimization (GRPO)~\cite{
shao2024deepseekmath},
driven by multi-seed execution rewards that account for executability, rule
satisfaction, structural diversity, scene integrity, and conjunct coverage.
An Understanding Replay objective is retained during policy refinement to
mitigate degradation in visual rule understanding.

We evaluate RuleForge across four multimodal backbones: UniLIP, InternVL3.5,
Qwen3-VL, and Show-o2~\cite{
tang2026unilip,
wang2025internvl3,
bai2025qwen3,
xie2026show}.
We additionally report zero-shot results from two commercial-model references,
GPT-6-Sol and Gemini-2.5 Flash-Lite~\cite{
openai2026gpt6sol,
google2025gemini25flashlite}.
On visual rule understanding, Stage~III UniLIP improves multiple-choice
accuracy over its base model by 20.93 percentage points on familiar rules,
from 64.08\% to 85.01\%, and by 28.42 percentage points on held-out compositions,
from 37.58\% to 66.00\%. On executable generation evaluated under separately
provided target-rule descriptions, its scene-level rule satisfaction (RuleSat)
increases from 2.91\% to 70.99\% on familiar rules and reaches 38.40\% on
held-out compositions. The other three backbones also improve over their
respective base models in multiple-choice accuracy and familiar-rule scene-level
RuleSat. Our main contributions are threefold:
\begin{itemize}[leftmargin=*]
    \item We formulate visual rule learning through the complementary
    evaluation and joint training of visual rule understanding and
    rule-conditioned program synthesis from separately provided target rules,
    and construct Bongard-Forge100k, a controlled diagnostic benchmark
    containing 108{,}797 problems with structured rule specifications,
    executable scene generators, and deterministic checkers.

    \item We propose RuleForge, a backbone-agnostic three-stage framework that
    combines progressive multimodal instruction tuning, multi-seed execution
    feedback, and Understanding Replay.

    \item We conduct systematic evaluations across four multimodal backbones,
    demonstrating consistent gains in visual rule understanding and executable
    program synthesis, and additionally report two zero-shot commercial-model
    references.
\end{itemize}

\section{Related Work}
\label{sec:related_work}

\noindent\textbf{Bongard Problems and Visual Rule Reasoning.}
Bongard problems study concept induction from small sets of positive and
negative visual examples~\cite{mikhail1970bongard}. Modern benchmarks include
programmatically generated geometric concepts in Bongard-LOGO~\cite{
nie2020bongard}, relational human--object interactions in Bongard-HOI~\cite{
jiang2022bongard}, and free-form concepts in real-world images in
Bongard-OpenWorld~\cite{wu2024bongard}. Recent analyses \cite{malkinski2024reasoning,depeweg2024solving} further show that
multimodal large language models still struggle with abstract rule induction \cite{wust2024bongard,jiang2024marvel}
and systematic visual reasoning~\cite{li2025seeing,wang2025glfomr}.
Existing evaluation protocols primarily assess query classification or
natural-language rule induction, leaving open whether a rule can also support
the construction of new valid instances. Langenfeld et al.~\cite{
langenfeld2026bongards} instead generate parameterized programmatic hypotheses
and fit them to classify candidate images. Bongard-Forge100k provides a
complementary generative setting in which models are evaluated on visual rule
understanding and rule-conditioned scene-program synthesis. RuleForge trains a
shared backbone for both capabilities, while keeping their input interfaces
explicitly separated.

\begin{table*}[!t]
    \centering
    \caption{\textbf{Comparison with representative Bongard benchmarks.}
    Inventory entries follow the native definitions of the respective
    benchmarks and are therefore not directly equivalent measures of concept
    complexity. ``Executable generation'' indicates whether generating
    executable scene-construction programs constitutes an evaluated model task.
    Bongard-LOGO uses rendering programs to construct its benchmark, but its
    standard evaluation task is binary classification.}
    \label{tab:dataset_comparison}

    \begingroup
    \footnotesize
    \renewcommand{\arraystretch}{1.05}
    \setlength{\tabcolsep}{3.5pt}

    \resizebox{\textwidth}{!}{%
    \begin{tabular}{
        @{}
        >{\raggedright\arraybackslash}p{3.8cm}|
        >{\raggedright\arraybackslash}p{4.0cm}
        >{\raggedright\arraybackslash}p{3.3cm}
        >{\raggedright\arraybackslash}p{2.1cm}
        >{\raggedright\arraybackslash}p{1.8cm}
        >{\raggedright\arraybackslash}p{4.2cm}
        @{}
    }
        \toprule
        \textbf{Benchmark} &
        \textbf{Understanding task} &
        \textbf{Executable generation} &
        \textbf{Visual domain} &
        \textbf{\# Problems} &
        \textbf{Concept inventory} \\
        \midrule

        Original Bongard Problems~\cite{mikhail1970bongard} &
        Verbal rule induction &
        No &
        Diagrams &
        100 &
        Problem-specific visual concepts \\

        Bongard-LOGO~\cite{nie2020bongard} &
        Binary classification &
        No &
        Drawings &
        12,000 &
        627 shape categories; 25 abstract attributes \\
        
        Bongard-HOI~\cite{jiang2022bongard} &
        Binary classification &
        No &
        HOI images &
        ${\sim}53{,}000$ &
        242 HOI concepts \\
        
        Bongard-OpenWorld~\cite{wu2024bongard} &
        Binary classification &
        No &
        Web images &
        1,010 &
        1,010 free-form concepts \\

        \textbf{Bongard-Forge100k} &
        Multiple-choice rule recognition; free-form rule explanation &
        \textbf{Yes (scene programs)} &
        Rendered scenes &
        \textbf{108,797} &
        62 registered base rules; 24 held-out compositions \\

        \bottomrule
    \end{tabular}%
    }

    \endgroup
\end{table*}

\noindent\textbf{Structured Visual Content and Unified Multimodal Models.}
Charts, infographics, and template-driven layouts illustrate how structured
visual content depends on quantitative, spatial, and layout relations rather
than isolated object appearance. ChartQA~\cite{masry2022chartqa} and
InfographicVQA~\cite{mathew2022infographicvqa} evaluate reasoning over such
content, while LayoutPrompter~\cite{lin2023layoutprompter} formulates layout
generation through explicit element and spatial constraints. Unified
multimodal models such as Janus~\cite{wu2025janus}, UniLIP~\cite{
tang2026unilip}, and Show-o2~\cite{xie2026show} further support combinations
of understanding and generation within shared architectures \cite{bachmann20244m}. RuleForge focuses
on whether the same backbone can acquire visual rule understanding and
rule-conditioned structured generation. Generated programs are executed in a
controlled scene environment and evaluated by deterministic rule checkers,
allowing relational constraints to be audited directly \cite{kanetou2026img2kg} rather than through
output plausibility alone.

\noindent\textbf{Program Synthesis and Execution-Grounded Learning.}
Executable programs provide intermediate representations for reasoning and
interfaces to external tools. PAL~\cite{gao2023pal} delegates symbolic
computation to a Python interpreter, whereas VisProg~\cite{gupta2023visual}
and ViperGPT~\cite{suris2023vipergpt} compose vision models and code modules
for multimodal queries. Self-Debugging~\cite{chen2024teaching} and
CRITIC~\cite{gou2024critic} use execution or tool feedback to revise outputs
at inference time. Training-time approaches \cite{jiang2024ledex} such as CodeRL~\cite{le2022coderl}
optimize code generation using functional correctness, while
DeepSeekMath~\cite{shao2024deepseekmath} introduces Group Relative Policy
Optimization for reasoning outputs. RuleForge extends execution-grounded
learning to parameterized visual-scene construction: candidate programs are
executed under multiple random seeds and evaluated for executability, rule
satisfaction, structural diversity, scene integrity, and conjunct coverage.
Understanding Replay, motivated by prior work that retains original objectives
during policy optimization~\cite{ouyang2022training}, limits degradation in
visual rule understanding \cite{zhang2024wings} while executable generation is refined.
% ============================================================
% Section 3: The RuleForge Benchmark
% ============================================================
\section{The Bongard-Forge100k Benchmark}
\label{sec:dataset}

Bongard-Forge100k evaluates two complementary capabilities: inferring a latent
visual rule from positive and negative examples, and constructing new scenes
that satisfy a given rule through executable programs. We use \emph{scene} to
denote a structured object configuration, \emph{panel} for its rendered image,
and \emph{problem} for an instance containing six positive and six negative
panels.

\begin{figure*}[t]
    \centering
    \includegraphics[width=\textwidth]{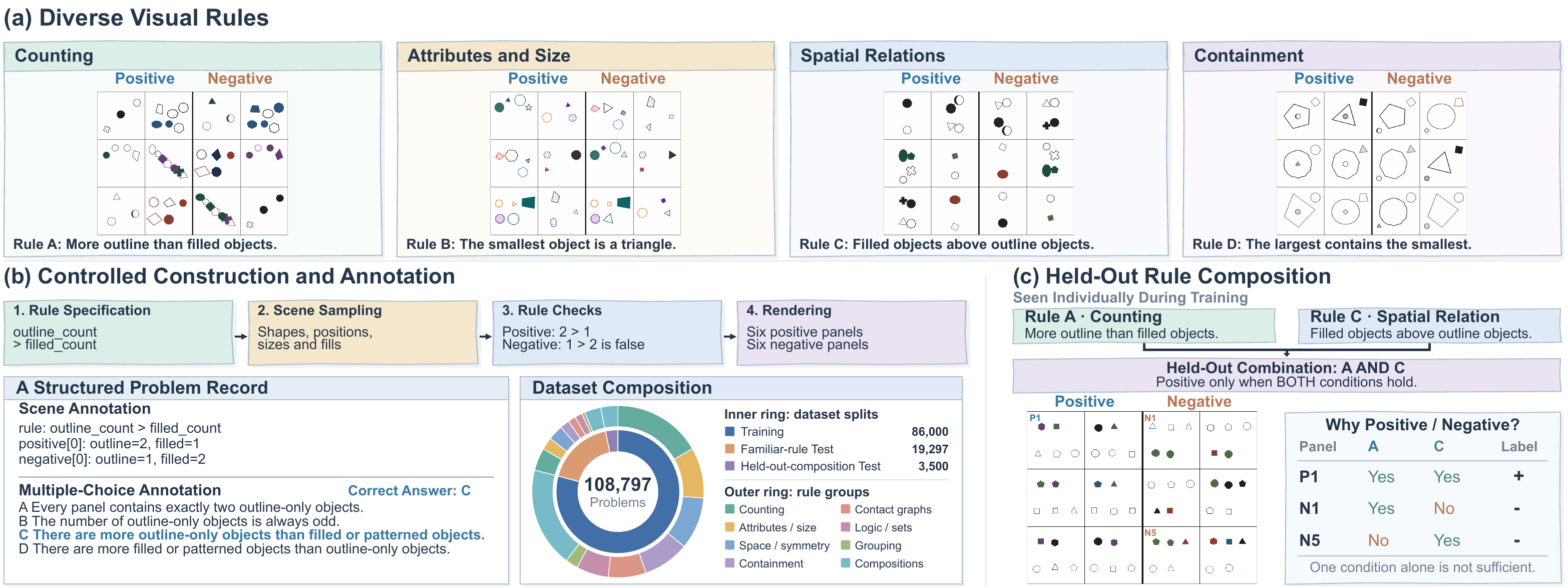}
    \vspace{-.2in}
    \caption{\textbf{Overview of Bongard-Forge100k.}
    (a) Representative problems covering counting, attributes and size,
    spatial relations, and containment.
    (b) The controlled construction pipeline, a structured problem record,
    and dataset composition. The donut chart's inner ring represents the
    training, familiar-rule test, and held-out-composition test splits. The
    outer ring partitions each split into eight display groups aggregated
    from finer rule-family metadata for legibility; sector sizes are
    proportional to problem counts.
    (c) Held-out compositions combine rules observed individually during
    training, requiring both constituent conditions to hold in positive
    examples.}
    \label{fig:dataset_overview}
    \vspace{-.1in}
\end{figure*}

We position Bongard-Forge100k relative to representative Bongard benchmarks
in Table~\ref{tab:dataset_comparison}. While prior benchmarks primarily
evaluate query classification or verbal rule induction, Bongard-Forge100k
additionally supports rule-conditioned executable scene-program generation.
Its construction pipeline is summarized in
Figure~\ref{fig:dataset_overview}.

\subsection{Rule Inventory and Specification}
\label{sec:dataset_rules}

The rule inventory draws on recurring concepts in classical Bongard
problems~\cite{mikhail1970bongard}, Bongard-LOGO~\cite{nie2020bongard},
and diagnostic visual reasoning datasets such as CLEVR~\cite{
johnson2017clevr}. We reorganize and extend these concepts into 62 registered
base rules across 15 semantic families, covering counting, attributes,
ranking, containment, topology, symmetry, alignment, grouping, contact
graphs, logical equivalence, and set structure. A base rule denotes a
registered construction unit and may contain more than one logical predicate.

Each rule $\varphi$ has four mutually consistent components: a
natural-language description $r_{\varphi}$, a structured specification, a
scene sampler, and a deterministic checker
$\chi_{\varphi}(S)\in\{0,1\}$. The checker determines whether scene $S$
satisfies the rule, making problem construction and generated-scene
evaluation auditable. The eight display groups in
Figure~\ref{fig:dataset_overview} aggregate finer rule-family metadata only
for visualization and do not alter the taxonomy of 62 base rules and
15 semantic families.

A rule can be instantiated through many combinations of object identities,
attributes, positions, and relations. Repeated problems therefore test
invariance to non-target visual changes, while held-out compositions test
whether familiar predicates can be recombined under unseen bindings.
Language models were used only to help draft some initial rule descriptions;
ground-truth semantics and labels are fixed by the structured specifications,
samplers, and deterministic checkers.

\subsection{Problem Construction and Supervision}
\label{sec:dataset_construction}
\label{sec:dataset_supervision}

For a target rule $\varphi$, the generator retains positive and negative
scenes satisfying
\begin{equation}
    \chi_{\varphi}(S_i^{+})=1,
    \qquad
    \chi_{\varphi}(S_i^{-})=0.
    \label{eq:dataset_scene_labels}
\end{equation}
Both groups share the same rendering vocabulary while varying in object
identity, color, fill, size, position, and other non-target properties.
Geometry and diversity checks reject malformed scenes, invalid relations,
out-of-canvas objects, and near duplicates.

For a conjunctive rule $\varphi=\varphi_1\land\varphi_2$, positive scenes
satisfy both constituents. Close negative counterexamples satisfy one
constituent while violating the other, preventing either constituent alone
from separating the groups. The generator also rejects problem sets in which
predefined non-target statistics, such as object counts or individual
attribute frequencies, perfectly separate the two groups.

Each accepted problem contains six positive and six negative scenes rendered
as a $4\times3$ contact sheet. Following the Bongard convention~\cite{
mikhail1970bongard,
nie2020bongard}, the positive and negative groups occupy the left and right
halves, respectively, while panels are shuffled within each group.

\noindent\textbf{Understanding supervision.}
Each visual problem supports multiple-choice rule recognition and free-form
rule explanation, together with auxiliary verification and factual tasks used
during training. The correct answer is derived from the registered rule and
structured annotation. For expanded multiple-choice problems, Qwen3.5-9B
proposes plausible distractors, which are screened for correctness,
distinctness, and support from the displayed panels. This model-assisted
process constructs challenging options but does not assign ground-truth
labels.

\noindent\textbf{Test-set fidelity and human audit.}
Automated checks verify scene polarity, annotations, option consistency, and
duplicate content. We also manually audited 1{,}000 randomly sampled test
problems. Of these, 966 passed all checks initially; the remaining 34 contained
an ambiguous or invalid option and were corrected and rechecked before
evaluation. The finalized benchmark retains its rule specifications,
generation provenance, audit status, and checker outputs.

\noindent\textbf{Executable-generation supervision.}
Generation records are maintained separately from visual problems. Each record
provides a target-rule description and the scene API as input, with a reference
scene program as output. During execution-grounded reinforcement learning, the
reference program is removed, and generated scenes are evaluated against the
structured target rule. Program records are therefore supervision views of the
shared rule system rather than additional Bongard-Forge100k problems.

\subsection{Dataset Statistics and Evaluation Splits}
\label{sec:dataset_statistics}

Bongard-Forge100k contains 108{,}797 visual reasoning problems: 86{,}000 for
training, 19{,}297 for familiar-rule testing, and 3{,}500 for testing
held-out rule compositions. The training set comprises 78{,}000 problems from
57 registered base rules and 8{,}000 problems from 40 training compositions.

The training data yield 172{,}000 understanding instruction records, including
rule explanation, multiple-choice recognition, candidate verification, and
factual prediction. Separate generation supervision provides 24{,}000 program
records covering all 62 base rules, while execution-grounded reinforcement
learning uses 16{,}000 prompts covering 163 rule compositions. These records do
not increase the count of independent visual problems.

The familiar-rule test set contains new visual instantiations of rules present
during training. The held-out set contains 3{,}500 problems from 24 predefined
rule compositions whose exact combinations are excluded from training. Thus,
24 denotes distinct held-out compositions, whereas 3{,}500 denotes their
visual instantiations under different object configurations and nuisance
attributes.

Executable-generation evaluation uses 223 familiar-rule prompts and one prompt
for each of the 24 held-out compositions. Each generated program is executed
under five fixed random seeds to measure executability and rule consistency
under scene variation. Unsuccessful executions remain part of the evaluation
rather than being discarded.

Before evaluation, the test sets are frozen using checks for
task--annotation correspondence, image decoding, exact and normalized-image
duplication, source overlap, recomputed rule measurements, and
multiple-choice consistency. Duplicate source or image content is excluded,
and source hashes are retained to make the finalized test sets auditable.
\section{Method}
\label{sec:method}
\label{sec:method_overview}

\subsection{Task Formulation and Unified Model}
\label{sec:task_formulation}

RuleForge trains a unified vision-language model (VLM) for two
complementary tasks that use different input interfaces but share the
same multimodal backbone and language decoder.

\noindent\textbf{Visual rule understanding.}
Given a contact sheet $I$ containing positive and negative examples
and an instruction $q_t$, the model predicts a response $y$. We
consider four output types: multiple-choice rule prediction,
natural-language explanation, candidate-rule verification, and
structured scene-fact prediction. These tasks respectively provide
supervision for rule discrimination, verbalization, consistency
checking, and visual grounding.

For an understanding record $(I,q_t,y)$ with target tokens
$y=(y_1,\ldots,y_{L_y})$, we optimize the target-token-normalized
understanding loss
\begin{equation}
    \mathcal{L}_{\mathrm{U}}
    =
    -\mathbb{E}_{(I,q_t,y)}
    \left[
    \frac{1}{L_y}
    \sum_{j=1}^{L_y}
    \log p_{\theta}
    \bigl(y_j\mid I,q_t,y_{<j}\bigr)
    \right].
\label{eq:understanding_loss}
\end{equation}
Only response tokens contribute to the loss. Length normalization
prevents longer explanations and structured fact sequences from
dominating shorter classification or verification responses.

\noindent\textbf{Rule-conditioned executable generation.}
Given a separately provided target-rule description $r_{\varphi}$ and
the scene-construction API $\mathcal{A}$, the model predicts a program
$c=(c_1,\ldots,c_{L_c})$ using
\begin{equation}
    \mathcal{L}_{\mathrm{G}}
    =
    -\mathbb{E}_{(r_{\varphi},\mathcal{A},c)}
    \left[
    \frac{1}{L_c}
    \sum_{j=1}^{L_c}
    \log p_{\theta}
    \bigl(c_j\mid r_{\varphi},\mathcal{A},c_{<j}\bigr)
    \right].
\label{eq:generation_loss}
\end{equation}
The program constructs positive scenes satisfying the target rule and
negative scenes violating it while allowing non-target properties to
vary. Execution makes the model's interpretation directly testable by
the deterministic Rule Checker.

The two tasks capture complementary directions of rule learning.
Understanding maps visual evidence into rule language, whereas
generation maps a separately provided rule into executable form.
Generation does not directly consume the contact sheet, separating
rule-recovery errors from program-synthesis errors while allowing both
capabilities to benefit from shared model parameters.

\begin{figure*}[t]
    \centering
    \includegraphics[width=\textwidth]{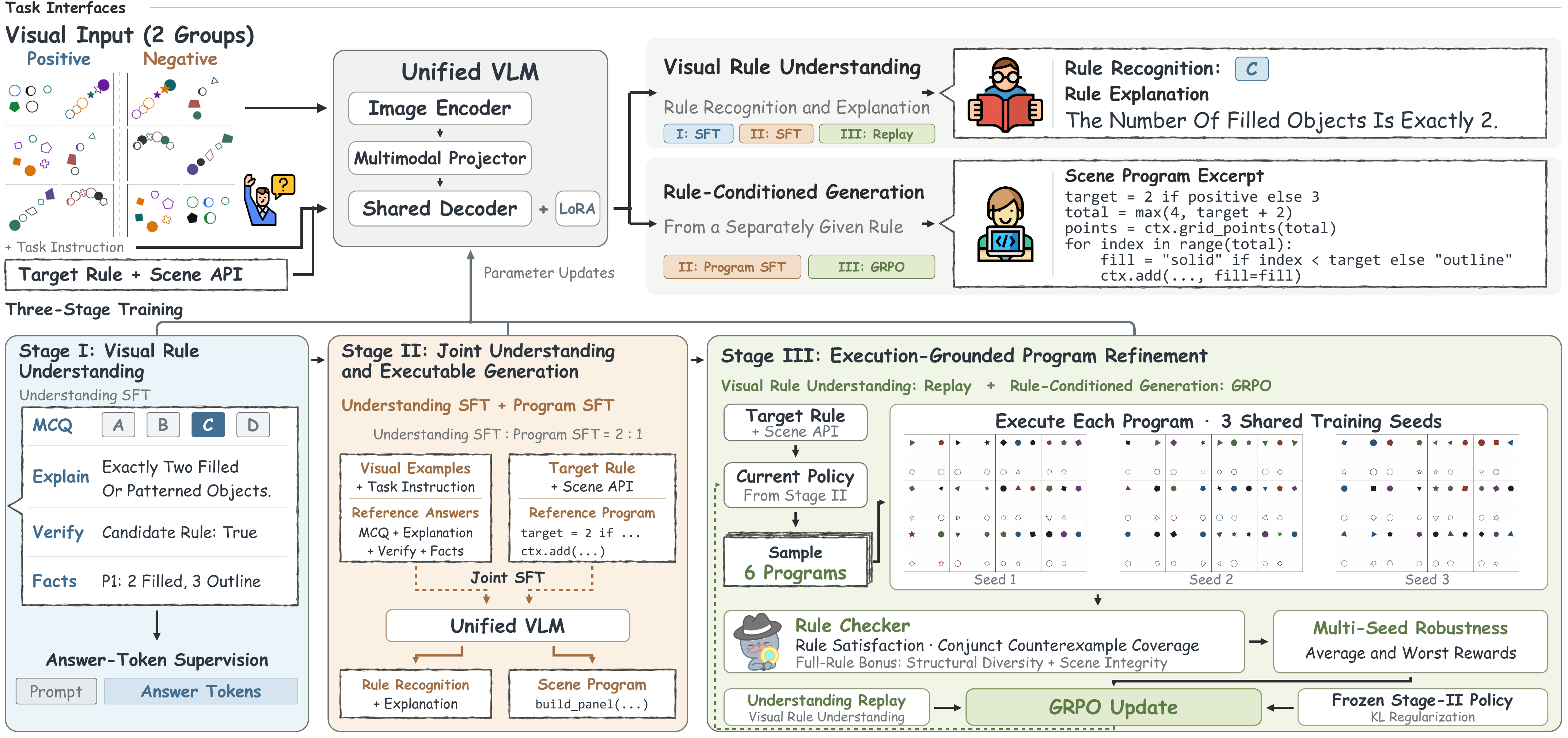}
    \caption{\textbf{Overview of RuleForge.}
    A unified VLM supports visual rule understanding from examples and
    rule-conditioned program generation from a separately given rule.
    Stage~I trains rule understanding; Stage~II jointly performs
    Understanding SFT and Program SFT; and Stage~III applies multi-seed
    execution feedback through GRPO with understanding replay. Six
    candidate programs are executed under three shared training seeds,
    while final evaluation uses five fixed seeds. The shared connector
    indicates that the same VLM is progressively updated across all
    stages.}
    \label{fig:method_overview}
\end{figure*}

Figure~\ref{fig:method_overview} summarizes the two interfaces and
three-stage training process. We abbreviate the generation interface
as \emph{Rule-Conditioned Generation}, the scene-construction API as
the \emph{Scene API}, and its supervision as \emph{Program SFT}.

We use LoRA~\cite{hu2021lora} for parameter-efficient adaptation. The
image encoder and base language-model weights remain frozen. Stage~I
updates the multimodal projector and language-model LoRA parameters,
whereas Stages~II and III freeze the projector and update only LoRA.
This preserves the visual alignment established in Stage~I while
allowing later stages to introduce executable generation and
execution-grounded refinement.

\subsection{Progressive Supervised Training}
\label{sec:supervised_training}

RuleForge first establishes visual rule understanding and then adds
program generation, providing the shared decoder with a stable
rule-oriented representation before it learns structured programs.

\noindent\textbf{Stage I: Visual Rule Understanding.}
Stage~I optimizes Equation~\ref{eq:understanding_loss} over the four
understanding tasks using a fixed task mixture. Multiple-choice
prediction supplies the main discriminative signal, while explanation,
verification, and scene-fact prediction provide complementary semantic
and perceptual supervision. The sampling probabilities are provided in
the appendix.

This multi-task design reduces dependence on any single response
format. Explanation requires independent rule verbalization,
verification evaluates a proposed rule against all examples, and
scene-fact prediction grounds the decision in object properties and
relations. Target-token masking and response-length normalization
balance short decisions against longer explanations and structured
outputs.

\noindent\textbf{Stage II: Joint Understanding and Executable
Generation.}
Stage~II retains Understanding SFT and introduces Program SFT using
Equation~\ref{eq:generation_loss}. Each Program-SFT record contains a
target-rule description, the Scene API, and a reference program. The
program produces six positive scenes satisfying the target rule and
six negative scenes violating it while varying rule-irrelevant
attributes. These records teach the syntax and composition patterns of
the Scene API without reducing generation to a fixed scene template.

Training interleaves two Understanding-SFT batches with one
Program-SFT batch. This $2{:}1$ schedule preserves visual rule
understanding while adding executable generation through the shared
decoder. We compare understanding-only, generation-only, $1{:}1$, and
$2{:}1$ configurations in Table~\ref{tab:ablation_results}(a), where
the selected mixture provides the strongest overall balance.

Program SFT teaches the model to imitate reference programs, but it
does not directly determine whether sampled programs execute safely,
satisfy the target rule, or remain reliable across random scene
realizations. Stage~III therefore evaluates generated programs through
execution and uses their observed behavior as the optimization signal.
\subsection{Execution-Grounded Program Refinement}
\label{sec:stage_c}

Stage~III applies GRPO~\cite{shao2024deepseekmath} to the generation
capability acquired in Stage~II. For each target rule, the current
model samples $K=6$ candidate programs. Every candidate is executed
under the same training seed set $\mathcal{G}_{\mathrm{tr}}$,
containing three seeds. Shared seeds control unrelated scene variation
during group-relative comparison. Unsafe programs, syntax failures,
and runtime failures receive failure-specific penalties. Final
generation metrics are recomputed using five fixed evaluation seeds
per program, as described in
Section~\ref{sec:experimental_setup}. Execution constraints and
failure penalties are provided in the appendix.

For every successful execution, the deterministic Rule Checker
evaluates six positive and six negative scenes. Positive scenes should
satisfy the target rule, whereas negative scenes should violate it.
The reward also considers structural diversity, scene integrity, and
conjunct counterexample coverage. Diversity and integrity contribute
only after full rule satisfaction, preventing visually varied but
semantically invalid programs from receiving additional credit. For
composite rules, conjunct coverage checks whether negative scenes
expose violations of different rule constituents.

To reduce dependence on favorable random choices, RuleForge considers
both average and worst-seed performance across the shared training
seeds. We evaluate the worst-seed and conjunct-coverage components in
Table~\ref{tab:ablation_results}(b). The full reward definition,
aggregation rule, and coefficients are provided in the appendix.

GRPO compares candidates within each group and updates the model
toward programs with stronger relative execution outcomes. We follow
the clipped policy-optimization procedure of GRPO and
PPO~\cite{schulman2017proximal}, using the frozen Stage~II model as the
reference policy. To prevent execution optimization from degrading
visual rule understanding, Stage~III also replays records from the
Stage~I task mixture using
Equation~\ref{eq:understanding_loss}. The optimization and
regularization settings are provided in the appendix. Understanding
Replay preserves rule recognition and explanation while execution
feedback improves rule-conditioned program generation.
\section{Experiment}
\label{sec:experiments}

\textbf{Experimental Setup.}\label{sec:experimental_setup}
We evaluate RuleForge on UniLIP-3B~\cite{tang2026unilip},
InternVL3.5-2B~\cite{wang2025internvl3},
Qwen3-VL-4B-Instruct~\cite{bai2025qwen3}, and
Show-o2-1.5B~\cite{xie2026show}. For each backbone, we compare the
base model with Stages~I--III. GPT-6-Sol and Gemini-2.5
Flash-Lite~\cite{google2025gemini25flashlite} serve as zero-shot
commercial references. All trainable models are trained once with
model-training seed 42 on a single NVIDIA GeForce RTX 4090 GPU. We
set $\beta_{\mathrm{KL}}=0.03$ and
$\lambda_{\mathrm{rep}}=1.0$ in Stage~III. Complete LoRA,
optimization, batching, decoding, and runtime settings are provided
in Appendix~\ref{app:implementation}. Visual rule understanding is evaluated on 19{,}297 familiar-rule and
3{,}500 held-out-composition Bongard-Forge100k problems.
Multiple-choice accuracy (MCQ) is computed by exact matching against
the ground-truth option. Explanation correctness (Exp.) is assessed
by a Qwen3.5-2B~\cite{qwen3.5} judge \cite{xiong2025lllava,chen2024mllm} under a fixed semantic-matching
rubric, and joint accuracy requires both outputs to be correct for the
same problem. Unparsable predictions and invalid judge outputs are
counted as incorrect. On 200 familiar-rule and 200 held-out-composition
samples, the judge achieves 94\% agreement with human assessment.

Executable generation is evaluated on 223 familiar-rule descriptions
and all 24 held-out compositions. Each program is executed under five
fixed seeds and produces six positive and six negative scenes per
seed. We report program executability (Exec.) and scene-level rule
satisfaction (RuleSat); unexecutable programs receive zero RuleSat.
Commercial models receive the full API specification as zero-shot
references, whereas trained models use short instructions after
supervised API learning. These results therefore do not constitute a
strictly matched-prompt comparison.

\subsection{Visual Rule Understanding}
\label{sec:understanding_results}

We report visual rule understanding results in
Table~\ref{tab:understanding_results}. On UniLIP, Stage~III raises
familiar-rule MCQ accuracy from 64.08\% to 85.01\% and held-out MCQ
accuracy from 37.58\% to 66.00\%. Stage~II gives the strongest UniLIP
explanation results, while the small Stage~II-to-III changes show that
execution-grounded refinement largely preserves rule understanding.

The gains transfer across architectures: the best familiar-rule MCQ
accuracy reaches 81.67\% for Show-o2, 87.29\% for Qwen3-VL, and
80.00\% for InternVL3.5. Explanation remains substantially harder
under held-out compositions. For example, GPT-6-Sol obtains 84.09\%
held-out MCQ accuracy but only 5.31\% explanation accuracy, whereas
Stage~III UniLIP reaches 16.46\% explanation accuracy. This gap shows
that selecting a candidate rule does not imply independently recovering
and verbalizing it.

\begin{table}[t]
\centering
\caption{\textbf{Visual rule understanding results (\%).}
MCQ, Exp., and Joint denote multiple-choice, explanation, and joint
accuracy. Bold marks the best RuleForge-trained result.}
\label{tab:understanding_results}
\setlength{\tabcolsep}{2.2pt}
\renewcommand{\arraystretch}{1.03}
\resizebox{\columnwidth}{!}{%
\begin{tabular}{@{}llrrrrrr@{}}
\toprule
\multirow{2}{*}{Model}
& \multirow{2}{*}{Stage}
& \multicolumn{3}{c}{Familiar Rules}
& \multicolumn{3}{c}{Held-Out Rule Compositions} \\
\cmidrule(lr){3-5}
\cmidrule(lr){6-8}
& & MCQ & Exp. & Joint & MCQ & Exp. & Joint \\
\midrule

\multicolumn{8}{@{}l}{\textbf{Commercial zero-shot references}} \\[1pt]

Gemini-2.5 Flash-Lite~\cite{google2025gemini25flashlite}
& -- & 30.95 & 10.37 & 3.61
& 37.63 & 0.74 & 0.23 \\

GPT-6-Sol
& -- & 68.81 & 21.32 & 16.18
& 84.09 & 5.31 & 4.63 \\

\addlinespace[4pt]
\multicolumn{8}{@{}l}{\textbf{Backbones trained with RuleForge}} \\[1pt]

\multirow{4}{*}{InternVL3.5-2B~\cite{wang2025internvl3}}
& Base & 33.31 & 3.81 & 1.61
& 35.00 & 0.97 & 0.76 \\
& I & 77.19 & 17.83 & 13.57
& 64.34 & 3.54 & 2.06 \\
& II & 79.92 & 18.82 & 15.15
& 67.46 & 1.06 & 0.91 \\
& III & 80.00 & 19.98 & 16.08
& 67.14 & 1.57 & 1.03 \\

\addlinespace[2pt]
\multirow{4}{*}{Show-o2-1.5B~\cite{xie2026show}}
& Base & 25.00 & 2.97 & 0.81
& 25.03 & 6.69 & 1.66 \\
& I & 75.63 & 12.66 & 9.91
& 41.29 & 1.14 & 0.57 \\
& II & 81.40 & 24.11 & 20.60
& 42.43 & 1.43 & 0.77 \\
& III & 81.67 & 22.61 & 19.13
& 42.31 & 1.43 & 0.86 \\

\addlinespace[2pt]
\multirow{4}{*}{UniLIP-3B~\cite{tang2026unilip}}
& Base & 64.08 & 4.78 & 2.90
& 37.58 & 0.23 & 0.15 \\
& I & 82.10 & 11.48 & 9.34
& \textbf{69.46} & 2.69 & 2.06 \\
& II & 84.95 & 25.80 & 22.55
& 66.60 & \textbf{16.83} & \textbf{7.26} \\
& III & 85.01 & 24.63 & 21.75
& 66.00 & 16.46 & 7.00 \\

\addlinespace[2pt]
\multirow{4}{*}{Qwen3-VL-4B-Instruct~\cite{bai2025qwen3}}
& Base & 42.32 & 17.62 & 7.15
& 62.51 & 0.69 & 0.57 \\
& I & 84.88 & 31.56 & 28.68
& 57.57 & 1.03 & 0.49 \\
& II & 87.15 & 38.30 & 35.25
& 58.69 & 1.29 & 0.69 \\
& III & \textbf{87.29} & \textbf{38.31} & \textbf{35.36}
& 58.71 & 1.71 & 0.80 \\

\bottomrule
\end{tabular}%
}
\vspace{-.10in}
\end{table}

\subsection{Rule-Conditioned Executable Generation}
\label{sec:generation_results}

In Table~\ref{tab:generation_results}, we evaluate whether a target-rule
description can be translated into an executable scene program.
The Base and Stage~I models rarely generate valid programs because
they receive no Program SFT. Stage~II establishes executable
generation across all four backbones, reaching 54.64--67.81\%
familiar-rule RuleSat. Stage~III further improves every backbone.
For UniLIP, familiar-rule RuleSat increases from 60.42\% to 70.99\%,
while held-out RuleSat rises from 23.33\% to 38.40\% and held-out
Exec.\ from 33.33\% to 54.17\%.

\begin{figure*}[!t]
    \centering
    \includegraphics[width=0.8\textwidth]{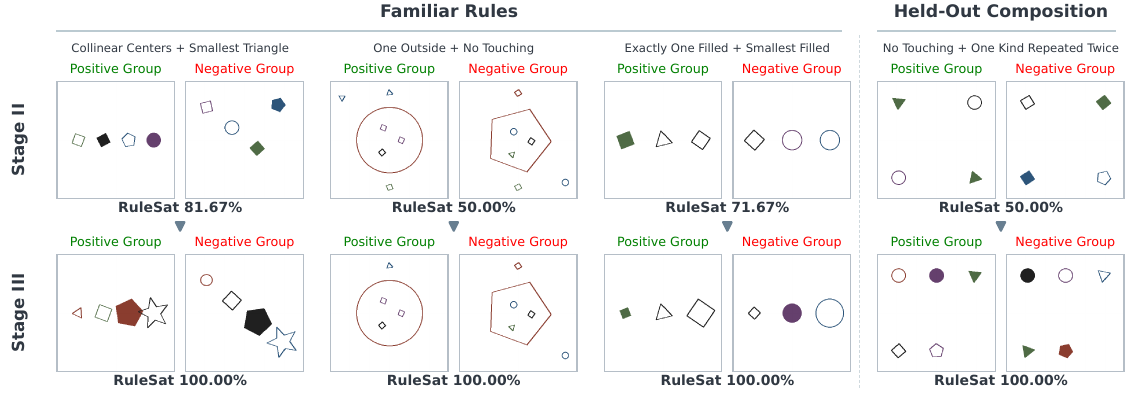}
    \vspace{-.1in}
    \caption{\textbf{Matched-seed comparison between Stages~II and III.}
    Stage~II produces executable but rule-violating positive scenes,
    whereas Stage~III satisfies the target rules. Each displayed case
    uses one seed; RuleSat is computed over five fixed seeds.}
    \label{fig:qualitative_results}
\end{figure*}

GPT-6-Sol achieves 85.41\% familiar-rule and 83.40\% held-out
RuleSat under the API-documented zero-shot setting, showing that a
strong general-purpose model can effectively use extensive API
documentation in context. Gemini-2.5 Flash-Lite reaches 30.37\% and
20.76\%, respectively, and is surpassed by every Stage~III backbone.
The consistent Stage~II-to-III gains provide the principal evidence
for execution-grounded refinement beyond supervised program
likelihood. In Figure~\ref{fig:qualitative_results}, we provide
matched-seed examples in which Stage~III corrects semantic rule
violations that remain after supervised Program SFT.

\begin{table}[t]
\centering
\caption{\textbf{Rule-conditioned executable generation results (\%).}
Exec.\ denotes program executability, and RuleSat denotes scene-level
rule satisfaction over five seeds. Bold marks the best
RuleForge-trained result.}
\label{tab:generation_results}
\setlength{\tabcolsep}{3.0pt}
\renewcommand{\arraystretch}{1.03}
\resizebox{\columnwidth}{!}{%
\small
\begin{tabular}{@{}llrrrr@{}}
\toprule
\multirow{2}{*}{Model}
& \multirow{2}{*}{Stage}
& \multicolumn{2}{c}{Familiar Rules}
& \multicolumn{2}{c}{Held-Out Rule Compositions} \\
\cmidrule(lr){3-4}
\cmidrule(lr){5-6}
& & Exec. & RuleSat & Exec. & RuleSat \\
\midrule

\multicolumn{6}{@{}l}{\textbf{API-documented commercial zero-shot references}} \\[1pt]

Gemini-2.5 Flash-Lite~\cite{google2025gemini25flashlite}
& -- & 45.47 & 30.37 & 33.33 & 20.76 \\

GPT-6-Sol
& -- & 94.71 & 85.41 & 91.67 & 83.40 \\

\addlinespace[4pt]
\multicolumn{6}{@{}l}{\textbf{Backbones trained with RuleForge}} \\[1pt]

\multirow{4}{*}{Qwen3-VL-4B-Instruct~\cite{bai2025qwen3}}
& Base & 4.04 & 2.69 & 0.00 & 0.00 \\
& I & 4.93 & 3.36 & 0.00 & 0.00 \\
& II & 69.24 & 54.64 & 37.50 & 25.00 \\
& III & 80.81 & 61.15 & 50.00 & 34.51 \\

\addlinespace[2pt]
\multirow{4}{*}{InternVL3.5-2B~\cite{wang2025internvl3}}
& Base & 2.24 & 1.46 & 0.00 & 0.00 \\
& I & 3.14 & 2.02 & 0.00 & 0.00 \\
& II & 75.78 & 59.57 & 29.17 & 18.75 \\
& III & 90.13 & 67.50 & 41.67 & 25.21 \\

\addlinespace[2pt]
\multirow{4}{*}{UniLIP-3B~\cite{tang2026unilip}}
& Base & 4.93 & 2.91 & 0.00 & 0.00 \\
& I & 4.93 & 2.69 & 0.00 & 0.00 \\
& II & 74.89 & 60.42 & 33.33 & 23.33 \\
& III & 87.44 & 70.99
& \textbf{54.17} & \textbf{38.40} \\

\addlinespace[2pt]
\multirow{4}{*}{Show-o2-1.5B~\cite{xie2026show}}
& Base & 3.59 & 2.02 & 0.00 & 0.00 \\
& I & 4.04 & 2.47 & 0.00 & 0.00 \\
& II & 83.86 & 67.81 & 12.50 & 10.42 \\
& III & \textbf{92.38} & \textbf{71.42}
& 41.67 & 22.99 \\

\bottomrule
\end{tabular}%
}
\end{table}

\subsection{Inference-Time Chaining through the Shared Rule-Language Interface}
\label{sec:experiments_chaining}

The preceding experiments evaluate understanding and generation under
their standard interfaces. To examine their inference-time connection,
we evaluate $I\rightarrow\hat{r}\rightarrow c$, where the model first
predicts a rule description $\hat{r}$ from the contact sheet and then
generates $c$ from
$p_{\theta}(c\mid\hat{r},\mathcal{A})$. We use matched,
rule-balanced subsets of 1{,}000 familiar-rule and 1{,}000
held-out-composition problems. For each Stage~II and Stage~III
checkpoint, \emph{Reference} supplies the ground-truth description,
whereas \emph{Predicted} uses the checkpoint's own free-form rule
prediction. Both conditions use the same API and five-seed execution
protocol. Sampling and bootstrap details are provided in
Appendix~\ref{app:chaining}.

\begin{table}[t]
\centering
\caption{\textbf{Inference-time rule chaining (\%).}
Ref.\ uses the reference rule; Pred.\ uses the model-predicted rule.
Bold indicates the better stage under the same rule source.}
\label{tab:predicted_rule_chaining}
\footnotesize
\setlength{\tabcolsep}{3pt}
\renewcommand{\arraystretch}{0.95}
\begin{tabular*}{0.95\columnwidth}{@{\extracolsep{\fill}}llrrrr@{}}
\toprule
& &
\multicolumn{2}{c}{Familiar} &
\multicolumn{2}{c}{Held-Out} \\
\cmidrule(lr){3-4}\cmidrule(lr){5-6}
Stage & Source
& Exec.\,$\uparrow$ & RuleSat\,$\uparrow$
& Exec.\,$\uparrow$ & RuleSat\,$\uparrow$ \\
\midrule
II  & Ref.  & 77.50 & 71.98 & 62.40 & 43.65 \\
III & Ref.  & \textbf{88.70} & \textbf{79.37} & \textbf{79.20} & \textbf{50.09} \\
\midrule
II  & Pred. & 59.20 & 34.43 & 25.60 & 12.96 \\
III & Pred. & \textbf{69.50} & \textbf{38.69} & \textbf{42.60} & \textbf{21.20} \\
\bottomrule
\end{tabular*}
\end{table}

In Table~\ref{tab:predicted_rule_chaining}, we observe a substantial
reference-to-predicted gap: under Stage~III, RuleSat decreases from
79.37\% to 38.69\% on familiar rules and from 50.09\% to 21.20\%
on held-out compositions. Nevertheless, Stage~III improves chained
RuleSat over Stage~II by 4.26 and 8.24 percentage points, with
rule-cluster bootstrap 95\% confidence intervals of $[0.99,7.44]$
and $[4.47,12.04]$, respectively; held-out Exec.\ also increases by
17.00 percentage points. These results demonstrate improvement in the
complete inference chain while identifying explicit rule recovery as a
major remaining bottleneck. Because each stage supplies both its own
descriptions and programs, the gains reflect their combined changes
rather than an isolated improvement in robustness to description
errors.

\subsection{Ablation Studies}
\label{sec:ablations}

We report the complete stage-wise results in
Tables~\ref{tab:understanding_results} and
\ref{tab:generation_results}, and visualize two representative trends
in Appendix Figure~\ref{fig:stage_progression}. We conduct controlled
UniLIP ablations of the Stage~II task mixture, the Stage~III execution
reward, and Understanding Replay.

\noindent\textbf{Stage-II task mixture.}
In Table~\ref{tab:ablation_results}(a), we compare understanding-only,
generation-only, and two joint mixtures. Understanding-only training
provides almost no executable generation, whereas generation-only
training reaches 68.76\% familiar RuleSat but substantially reduces
understanding accuracy. The $2{:}1$ mixture obtains the strongest
familiar and held-out MCQ while remaining competitive on explanation
and generation, motivating its use as the default.

\noindent\textbf{Stage-III execution reward.}
In Table~\ref{tab:ablation_results}(b), we remove the worst-seed or
conjunct coverage component. Removing the worst-seed term slightly
improves familiar RuleSat but reduces held-out RuleSat from 38.40\%
to 26.74\%. Removing coverage similarly weakens held-out RuleSat,
executability, worst-seed satisfaction, and coverage. The full reward
therefore primarily improves robustness under held-out compositions.

\begin{table}[t]
\centering
\caption{\textbf{UniLIP-3B ablations (\%).}
(a) Stage~II task mixtures; (b) Stage~III reward components.}
\label{tab:ablation_results}
\vspace{-.05in}

\setlength{\tabcolsep}{3.2pt}
\renewcommand{\arraystretch}{0.82}

\resizebox{0.75\columnwidth}{!}{%
\begin{tabular}{@{}lrrrr@{}}
\multicolumn{5}{c}{\textbf{(a) Stage~II task mixture}} \\[-1pt]
\toprule
Metric
& U-only
& G-only
& $1{:}1$
& $2{:}1$ \\
\midrule
Familiar MCQ
& 70.80 & 61.50 & 72.90 & \textbf{84.95} \\
Held-out MCQ
& 60.77 & 49.74 & 59.73 & \textbf{66.60} \\
Familiar Exp.
& 24.80 & 5.40 & \textbf{25.90} & 25.80 \\
Held-out Exp.
& 8.22 & 0.10 & \textbf{17.07} & 16.83 \\
Familiar RuleSat
& 2.91 & \textbf{68.76} & 61.89 & 60.42 \\
Held-out RuleSat
& 0.00 & 21.18 & \textbf{27.36} & 23.33 \\
\bottomrule
\end{tabular}%
}

\vspace{-.01in}

\resizebox{0.75\columnwidth}{!}{%
\begin{tabular}{@{}lrrr@{}}
\multicolumn{4}{c}{\textbf{(b) Stage~III execution reward}} \\[-1pt]
\toprule
Metric
& w/o Worst
& w/o Coverage
& Full \\
\midrule
Familiar RuleSat
& \textbf{71.24} & 70.06 & 70.99 \\
Held-out RuleSat
& 26.74 & 30.14 & \textbf{38.40} \\
Held-out Exec.
& 45.83 & 46.67 & \textbf{54.17} \\
Held-out WorstRS
& 25.69 & 29.17 & \textbf{36.46} \\
Held-out Coverage
& 17.50 & 16.11 & \textbf{19.58} \\
\bottomrule
\end{tabular}%
}

\vspace{-.08in}
\end{table}

\noindent\textbf{Understanding Replay.}
We ablate Understanding Replay by retraining Stage~III from the same
Stage~II checkpoint without $\mathcal{L}_{\mathrm{rep}}$.
As detailed in Appendix Table~\ref{tab:replay_ablation}, removing
replay increases held-out MCQ by 1.29 percentage points, but decreases
explanation, joint, executability, and RuleSat by 3.37, 1.40, 8.34,
and 8.05 points, respectively. These results support Understanding
Replay as a cross-task regularizer that preserves explicit rule
verbalization and executable generation, rather than as a uniform
accuracy booster.

\subsection{Qualitative Results}
\label{sec:qualitative_results}

Figure~\ref{fig:qualitative_results} presents matched-seed comparisons
between programs produced by Stages~II and III. Although Stage~II
generally produces executable and visually valid scenes, it may still
violate alignment, contact, attribute, counting, or conjunctive
constraints. Stage~III corrects these semantic violations while
preserving variation in rule-irrelevant properties such as object
identity, color, size, and position. The displayed familiar-rule cases
improve from Stage~II RuleSat values of 50.00--81.67\% to 100.00\%,
while the held-out example jointly satisfies both constituents of an
unseen composition. The examples also reveal that executability alone
is insufficient: syntactically valid programs can still encode
incomplete rule logic, whereas execution-grounded refinement improves
semantic consistency across multiple realizations and random seeds.
Because corresponding programs are rendered under the same seeds,
these gains cannot be attributed solely to favorable random
configurations. Instead, these cases show that execution feedback
changes the underlying construction logic, complementing the aggregate
results in Table~\ref{tab:generation_results} with concrete improvements
in generated scene structure.
\section{Conclusion}
\label{sec:conclusion}

We presented Bongard-Forge100k, a controlled benchmark for visual rule
understanding and rule-conditioned executable generation, together with
RuleForge, a backbone-agnostic framework for jointly learning these two
capabilities. Through progressive understanding training, joint
scene-program supervision, and execution-grounded reinforcement learning
with Understanding Replay, RuleForge improves both rule understanding and
executable generation across UniLIP, InternVL3.5, Qwen3-VL, and Show-o2.
The Stage III UniLIP model surpasses the zero-shot GPT-6-Sol and Gemini-2.5
Flash-Lite references on all familiar-rule understanding metrics. It also
surpasses both references on explanation and joint accuracy for held-out
compositions, demonstrating the value of task-specific training for explicit
rule recovery. Across the full training pipeline, familiar-rule RuleSat
increases from 2.91\% to 70.99\%. More specifically,
execution-grounded refinement improves RuleSat from 60.42\% to 70.99\%
on familiar rules and from 23.33\% to 38.40\% on held-out compositions,
supporting the effectiveness of multi-seed execution feedback. Together,
Bongard-Forge100k and RuleForge provide a controlled foundation for
learning and evaluating semantic and executable representations of visual
rules, with potential extensions to rule-governed multimodal Web content
such as interfaces, layouts, and infographics.

\bibliographystyle{ACM-Reference-Format}
\bibliography{references}

\clearpage
\appendix

\twocolumn[{%
  \begin{center}
    {\LARGE\bfseries
    RuleForge: Inferring and Instantiating Latent Visual Rules as
Executable Programs\par}
    \vspace{0.5em}
    {\Large\scshape Supplementary Material\par}
  \end{center}
  \vspace{0.5em}
}]

\section{Implementation and Evaluation Details}
\label{app:implementation}

\subsection{Parameter-Efficient Adaptation}

For all four backbones, the image encoder and base language-model
weights remain frozen throughout training. We insert LoRA adapters
into the language model's query, key, value, and output projections,
as well as its gate, up, and down projections. Unless otherwise
specified, LoRA uses rank 32, scaling factor 64, and dropout 0.05.
Stage~I additionally updates the multimodal projector; the projector
is frozen in Stages~II and III.

Training uses BF16 precision without weight quantization. We optimize
with AdamW, weight decay 0.01, a warmup ratio of 0.03, and gradient
clipping at 1.0. All reported trainable models use model-training seed
42 and are trained once on a single NVIDIA GeForce RTX 4090 GPU.
We summarize the stage-specific configuration in
Table~\ref{tab:training_configuration}.

\begin{table*}[t]
\centering
\caption{\textbf{Training configuration used for the three RuleForge
stages.}
Logical batch sizes are measured before gradient accumulation.
Stage~III samples one group of six candidate programs per policy
update.}
\label{tab:training_configuration}

\small
\setlength{\tabcolsep}{3.5pt}
\renewcommand{\arraystretch}{1.18}

\begin{tabularx}{\textwidth}{
    @{}
    c
    >{\raggedright\arraybackslash}p{2.7cm}
    >{\raggedright\arraybackslash}p{2.3cm}
    >{\raggedright\arraybackslash}p{2.5cm}
    >{\raggedright\arraybackslash}p{3.2cm}
    >{\raggedright\arraybackslash}X
    @{}
}
\toprule
\textbf{Stage}
& \textbf{Initialization and trainable components}
& \textbf{Training budget}
& \textbf{Learning rate}
& \textbf{Batching}
& \textbf{Sampling configuration} \\
\midrule

I
& Base model; LM LoRA and multimodal projector
& 2 epochs; 6,500 updates
& $1{\times}10^{-5}$ LoRA; $2{\times}10^{-6}$ projector
& Logical batch 8; accumulation 6; effective batch 48
& MCQ/Exp./Verify/Fact $=0.60/0.25/0.10/0.05$ \\
\hline
II
& Stage~I checkpoint; LM LoRA only
& 1 epoch; 7,313 updates
& $1{\times}10^{-5}$
& Understanding and program batches 8; accumulation 4
& Understanding:Program $=2{:}1$ \\
\hline
III
& Stage~II checkpoint; LM LoRA only
& 1,000 policy updates
& $1{\times}10^{-6}$
& Six candidates per group; replay batch 8
& Three shared execution seeds; understanding replay at every update \\

\bottomrule
\end{tabularx}
\end{table*}

\subsection{Stage-Specific Training Details}

\noindent\textbf{Stage~I.}
The understanding sampler draws from multiple-choice prediction,
free-form explanation, candidate verification, and structured
scene-fact prediction with probabilities 0.60, 0.25, 0.10, and 0.05,
respectively. Each epoch contains 156{,}000 sampled records, and the
model is trained for two epochs. The full training pool contains
172{,}000 understanding records. The sampler traverses shuffled
cycles while maintaining the prescribed task mixture and
rule-balanced sampling.

\noindent\textbf{Stage~II.}
Stage~II alternates two Understanding-SFT batches with one Program-SFT
batch. Both batch types contain eight records, and four logical batches
are accumulated before an optimizer update. Understanding responses
are capped at 1,024 tokens in the training sequence, while scene
programs use a maximum sequence length of 1,536 tokens. Stage~II is
trained for one epoch from the Stage~I checkpoint. The multimodal
projector remains frozen.

\noindent\textbf{Stage~III.}
For every target-rule prompt, the policy samples six candidate
programs with temperature 0.8, top-$p=1.0$, a maximum prompt length of
1,024 tokens, and at most 768 newly generated tokens. Candidates in
the same group are executed under the same three random seeds. The
GRPO clipping threshold is 0.2, and one policy epoch is applied per
rollout group. We set the worst-seed aggregation weight to 0.25, the
conjunct-coverage penalty coefficient to 0.25,
$\beta_{\mathrm{KL}}=0.03$, and
$\lambda_{\mathrm{rep}}=1.0$. Each update includes eight understanding
records drawn from the Stage~I task mixture. LoRA dropout is disabled
during on-policy rollout and optimization.

Unsafe programs, syntax failures, and runtime failures receive rewards
of $-1.0$, $-0.8$, and $-0.5$, respectively. Program execution occurs
in a restricted environment with a fixed API, resource limits, and
failure handling. Candidates that produce no valid execution do not
receive a positive policy reward.

\subsection{Input Processing, Decoding, and Runtime}

The complete Bongard contact sheet is resized using each backbone's
native image processor without cropping away boundary panels.
UniLIP, InternVL3.5, and Qwen3-VL use a $448\times448$ input, while
Show-o2 uses $432\times432$. Architecture-specific visual token
layouts and positional encodings are otherwise preserved.

Understanding evaluation uses deterministic decoding with maximum
output lengths of 32 new tokens for multiple-choice prediction and
192 new tokens for rule explanation. Program evaluation allows up to
1,536 new tokens. Every generated evaluation program is executed
under the same five fixed seeds:
\[
\{9100001,9100002,9100003,9100004,9100005\}.
\]
Each seed produces six positive and six negative scenes, yielding
60 structured scenes per evaluated program.

For an executable program, per-seed RuleSat is the fraction of its six
positive scenes accepted by the target-rule checker plus the fraction
of its six negative scenes rejected by the checker, averaged over all
12 scenes. The final score averages over the five seeds and evaluation
prompts. Unexecutable programs receive RuleSat zero and remain in the
denominator.

The primary UniLIP training pipeline requires approximately 39
wall-clock hours on one RTX 4090. Runtime varies across backbones
because their visual tokenizers, attention implementations, and
generation speeds differ. Training-seed 42 controls model
initialization and data sampling; it is distinct from the three
execution seeds used for Stage~III feedback and the five fixed seeds
used for final generation evaluation.

\section{Inference-Time Chaining Details}
\label{app:chaining}

The chaining experiment connects the two RuleForge interfaces without
introducing additional training. Given a Bongard contact sheet $I$,
the checkpoint first produces a free-form rule description
$\hat{r}$. The same checkpoint then receives $\hat{r}$ together with
the scene-construction API $\mathcal{A}$ and generates a program $c$:
\[
I \rightarrow \hat{r} \rightarrow c.
\]
The \emph{Reference} condition replaces $\hat{r}$ with the
ground-truth target-rule description while keeping the generation
prompt, API specification, decoding configuration, and execution
seeds unchanged. The \emph{Predicted} condition uses the checkpoint's
own rule description without manual correction. Missing or invalid
outputs remain failures rather than being removed.

We construct fixed, rule-balanced subsets of 1{,}000 familiar-rule
and 1{,}000 held-out-composition problems. Sampling quotas are
distributed as evenly as possible across rule identifiers, followed
by deterministic within-rule sampling. The same problem identifiers
are used for Stages~II and III and for the Reference and Predicted
conditions. Consequently, differences between stages are paired over
the same visual inputs and execution seeds.

Every generated program is evaluated under the five fixed seeds used
in the main generation experiment. The reported Exec.\ and RuleSat
metrics use the same failure and averaging rules as
Table~\ref{tab:generation_results}. The chaining results therefore
measure the complete inference path rather than a separately trained
image-to-code model.

Because multiple problems share the same underlying rule, individual
problems are not treated as statistically independent. We form
confidence intervals by resampling rule clusters with replacement,
including all selected problems associated with each sampled cluster,
and recomputing the paired Stage~III-minus-Stage~II difference. The
reported 95\% intervals are the 2.5th and 97.5th percentiles of the
resulting cluster-bootstrap distribution. This procedure gives
intervals of $[0.99,7.44]$ and $[4.47,12.04]$ percentage points for
the familiar-rule and held-out RuleSat gains, respectively.

% A single double-column float keeps the replay table and stage-wise
% visualization aligned at the top of the same page.
% Table 7: restore the original single-column size.
\begin{table}[t]
    \centering
    \caption{\textbf{Understanding Replay ablation on UniLIP-3B (\%).}
    The ablated Stage~III model removes $\mathcal{L}_{\mathrm{rep}}$.
    Bold indicates the better result in each column.}
    \label{tab:replay_ablation}

    \footnotesize
    \renewcommand{\arraystretch}{1.0}
    \setlength{\tabcolsep}{3.0pt}

    \textbf{(a) Visual Rule Understanding}
    \vspace{2pt}

    \begin{tabular}{@{}lrrrrrr@{}}
        \toprule
        &
        \multicolumn{3}{c}{Familiar} &
        \multicolumn{3}{c}{Held-Out} \\
        \cmidrule(lr){2-4}
        \cmidrule(l){5-7}
        Configuration &
        MCQ & Exp. & Joint &
        MCQ & Exp. & Joint \\
        \midrule
        Full III
        & \textbf{85.01} & 24.63 & 21.75
        & 66.00 & \textbf{16.46} & \textbf{7.00} \\
        III w/o Replay
        & 84.91 & \textbf{25.30} & \textbf{22.35}
        & \textbf{67.29} & 13.09 & 5.60 \\
        \bottomrule
    \end{tabular}

    \vspace{5pt}

    \textbf{(b) Rule-Conditioned Executable Generation}
    \vspace{2pt}

    \begin{tabular}{@{}lrrrr@{}}
        \toprule
        &
        \multicolumn{2}{c}{Familiar} &
        \multicolumn{2}{c}{Held-Out} \\
        \cmidrule(lr){2-3}
        \cmidrule(l){4-5}
        Configuration &
        Exec. & RuleSat &
        Exec. & RuleSat \\
        \midrule
        Full III
        & \textbf{87.44} & \textbf{70.99}
        & \textbf{54.17} & \textbf{38.40} \\
        III w/o Replay
        & 83.95 & 65.97
        & 45.83 & 30.35 \\
        \bottomrule
    \end{tabular}
\end{table}

% Figure 5: keep it as an independent single-column figure.
\begin{figure}[t]
    \centering
    \includegraphics[width=\columnwidth]{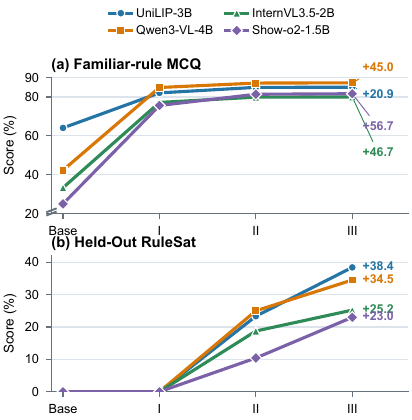}
    \caption{\textbf{Stage-wise performance.}
    (a) Familiar-rule MCQ accuracy and
    (b) held-out RuleSat for the base model and Stages~I--III.
    Stage~III annotations show percentage-point gains over the
    corresponding base models.}
    \label{fig:stage_progression}
\end{figure}

\section{Additional Stage-Wise and Ablation Results}
\label{app:additional_results}

\subsection{Stage-Wise Trends}

In Figure~\ref{fig:stage_progression}, we summarize how rule
understanding and executable generation emerge across the three
training stages. Stage~I produces the main improvement in familiar-rule
selection: MCQ accuracy rises from 64.08\% to 82.10\% for UniLIP, from
33.31\% to 77.19\% for InternVL3.5, from 42.32\% to 84.88\% for
Qwen3-VL, and from 25.00\% to 75.63\% for Show-o2. These gains occur
before program supervision, confirming that they arise from the
understanding-focused instruction mixture rather than executable
generation training.

Familiar-rule MCQ accuracy remains comparatively stable or improves
further during Stages~II and III. This behavior indicates that adding
Program SFT and execution-grounded optimization does not erase the
rule-selection ability acquired in Stage~I. The smaller changes after
Stage~I also suggest that later training primarily changes how the
model operationalizes a rule rather than whether it can select the
corresponding description from a candidate set.

The generation trajectory is qualitatively different. Held-out
RuleSat remains zero or near zero for the base and Stage~I models
because neither checkpoint has received scene-program supervision.
Stage~II establishes the executable-generation interface, reaching
held-out RuleSat values of 23.33\% for UniLIP, 25.00\% for Qwen3-VL,
18.75\% for InternVL3.5, and 10.42\% for Show-o2. Stage~III raises
these scores to 38.40\%, 34.51\%, 25.21\%, and 22.99\%,
respectively.

The consistent Stage~II-to-III gains are particularly relevant because
the exact held-out rule compositions do not occur in training.
Execution feedback therefore improves more than the syntactic
production of programs: it encourages programs whose behavior remains
consistent with the target rule under multiple scene realizations.
Together, the two panels show the intended division of labor across
the curriculum: Stage~I establishes visual rule understanding,
Stage~II introduces executable synthesis, and Stage~III improves
cross-seed rule satisfaction.

\subsection{Complete Understanding Replay Ablation}

We report the complete Understanding Replay ablation in
Table~\ref{tab:replay_ablation}. Replay is not a uniform accuracy
booster. Removing it changes all familiar-rule understanding metrics
by at most 0.67 percentage points and slightly improves familiar
explanation and joint accuracy. On held-out compositions, however,
explanation and joint accuracy decrease by 3.37 and 1.40 percentage
points, despite a 1.29-percentage-point increase in MCQ accuracy.

This difference between multiple-choice selection and free-form
explanation is informative. Candidate options provide explicit rule
descriptions that can support recognition even when the model's
ability to independently recover and verbalize the complete rule has
drifted. Explanation accuracy instead requires the model to express
all relevant predicates and their bindings without access to
candidate descriptions. Understanding Replay therefore has its
clearest semantic effect on the more demanding free-form interface.

The effect is more consistent for executable generation. Removing
replay reduces familiar Exec.\ and RuleSat by 3.49 and 5.02 points,
respectively, and reduces held-out Exec.\ and RuleSat by 8.34 and
8.05 points. The larger held-out decreases indicate that replay is
especially useful when program generation must preserve rule
information under an unseen composition rather than reproduce a
familiar rule interface.

These results support Understanding Replay as a cross-task
regularizer. It does not maximize every individual understanding
metric, but it limits semantic drift during policy optimization and
preserves information that remains useful for rule-conditioned
program generation. The replay objective therefore complements the
KL reference: the latter constrains the policy relative to Stage~II
on program tokens, whereas replay continues to expose the model to
the visual understanding tasks learned in Stage~I.

\end{document}